\documentclass{article}

\usepackage{microtype}
\usepackage{graphicx}
\usepackage{subcaption}
\usepackage{booktabs}
\usepackage{float}
\usepackage{tabularx}
\usepackage{array}
\usepackage{enumitem}
\usepackage{amsmath,amssymb,mathtools}
\usepackage{xcolor}
\usepackage{tikz}
\usetikzlibrary{arrows.meta,positioning,fit,shapes.geometric,backgrounds,calc}
\usepackage{hyperref}

\usepackage[accepted]{icml2026}

\newcommand{\methodname}{Pluralistic Lifecycle Governance}
\newcommand{\plg}{\textsc{PLG}}

\newcolumntype{Y}{>{\raggedright\arraybackslash}X}
\newcolumntype{L}[1]{>{\raggedright\arraybackslash}p{#1}}

\definecolor{deepblue}{RGB}{28,76,125}
\definecolor{lightblue}{RGB}{232,242,252}
\definecolor{lightgreen}{RGB}{236,248,239}
\definecolor{lightred}{RGB}{252,236,235}
\definecolor{lightgold}{RGB}{255,247,225}
\definecolor{darkgreen}{RGB}{35,105,61}
\definecolor{darkred}{RGB}{148,50,44}

\icmltitlerunning{AI Pluralism and the Worlds It Misses}

\begin{document}
\twocolumn[
  \icmltitle{AI Pluralism and the Worlds It Misses}

  \begin{icmlauthorlist}
    \icmlauthor{Rashid Mushkani}{udem,mila}
  \end{icmlauthorlist}

  \icmlaffiliation{udem}{Universit{\'e} de Montr{\'e}al, Montr{\'e}al, Qu{\'e}bec, Canada}
  \icmlaffiliation{mila}{Mila -- Qu{\'e}bec AI Institute, Montr{\'e}al, Qu{\'e}bec, Canada}

  \icmlcorrespondingauthor{Rashid Mushkani}{rashidmushkani@gmail.com}

  \icmlkeywords{pluralistic alignment, participatory AI, AI governance, ontological flattening, lifecycle accountability}

  \vskip 0.3in
]
\printAffiliationsAndNotice{}

\begin{abstract}
AI pluralism is often framed as a problem of representing diverse values, preferences, users, or outputs. This paper argues that this framing is incomplete because AI systems also impose ontologies: they define what counts as an entity, relation, feature, harm, benefit, and valid form of evidence. We define \emph{ontological flattening} as the conversion of situated, contested, and historically specific meanings into a restricted technical category, proxy, aggregation rule, or benchmark target that is treated as neutral and difficult to contest. The paper develops a bounded conceptual and qualitative synthesis across value pluralism, pluralistic alignment, participatory and democratic AI, procedural justice, science and technology studies, accountability research, aggregate themes from 11 expert interviews, and three urban AI companion cases. The cases illustrate how pluralistic methods can improve or structure model behavior while still compressing categories, proxies, aggregation rules, and revision rights before affected actors have procedural standing. We introduce \methodname{} (\plg{}) as a preliminary qualitative audit scaffold for documenting ontological openness, epistemic inclusion, procedural authority, evaluation pluralism, and lifecycle accountability. \plg{} is not presented as a validated scoring instrument; it is a framework for making the evidence and governance conditions of pluralistic AI explicit.
\end{abstract}

\section{Introduction}

AI systems do not only predict, rank, classify, or generate. They define the terms under which the world becomes computable. A street becomes a vector of visual features. A public-space rendering becomes a preference comparison. A community judgment becomes a benchmark label. A contested concept such as safety, accessibility, comfort, or inclusion becomes a category that a model can optimize. These translations are often necessary for computation, but they are not neutral. They decide which distinctions matter, which forms of knowledge count, and which conflicts remain visible.

Pluralistic alignment and participatory AI have become important responses to this problem. Recent work argues that AI systems should represent a wider range of values, expose multiple reasonable answers, support steering across perspectives, or match population distributions \citep{sorensen2024roadmap}. Normative accounts further show that pluralistic value alignment requires choices about criteria, origins, measurement, aggregation, and legitimacy \citep{kasirzadeh2024plurality}. These developments reject the assumption that alignment means optimizing one global preference function. Yet they leave a prior question underdeveloped: when do pluralistic AI methods preserve contestable ways of world-making, and when do they compress them into fixed technical ontologies?

This paper argues that AI pluralism must address ontological pluralism as well as value pluralism. By \emph{ontology}, we mean the practical representational commitments embedded in an AI system about what exists, what can be measured, how entities relate, and what counts as evidence. By \emph{lifeworld}, we mean the situated background of meanings, identities, histories, and practical judgments through which people inhabit social life. By \emph{procedural standing}, we mean the ability of affected actors to influence, contest, revise, or appeal decisions about a system rather than merely supply data to it. A public street can be enacted as infrastructure, memory, risk, care, commerce, surveillance, refuge, exclusion, or belonging. A benchmark that reduces this plurality to a stable label such as \emph{inclusive} or \emph{safe} may be useful, but it also performs a world-making act.

Habermas's account of lifeworld colonization provides one diagnostic vocabulary for this failure. He distinguishes communicatively reproduced lifeworlds from systems governed by money, power, and instrumental rationality \citep{habermas1984,habermas1987}. In AI, colonization appears when contested judgments are turned into prediction tasks, local knowledge becomes a feature vector, public reason becomes a benchmark score, and communities may contest outputs but not the representational scheme that produced them. This is not only bias, misclassification, or measurement error. It is \emph{ontological flattening}: the conversion of situated, contested, and historically specific worlds into restricted technical categories or proxies treated as neutral.

The empirical scope is deliberately bounded. We focus on urban, public-space, and vision-based systems because accessibility, safety, inclusion, comfort, and belonging are not stable visual properties. The evidence base combines a concept map of relevant literatures, 11 semi-structured expert interviews, and secondary analysis of three companion cases: Case A, a pluralistic public-space generation dataset and Direct Preference Optimization experiment; Case B, a participatory streetscape inclusivity model; and Case C, a reliability-aware urban vision-language benchmark \citep{anon2025casea,anon2026caseb,anon2025casec}. The cases support analytic generalization about lifecycle mechanisms, not statistical generalization across AI domains.

The paper makes three contributions. First, it defines ontological flattening as a failure mode that can persist even when a system supports pluralistic outputs or collects pluralistic feedback. Second, it distinguishes \emph{outcome pluralism}, where systems produce, steer, or approximate diverse outputs, from \emph{procedural pluralism}, where affected actors have standing over categories, evidence, aggregation, evaluation, and revision. Third, it proposes \methodname{} (\plg{}) as a qualitative framework for documenting and contesting ontological openness, epistemic inclusion, procedural authority, evaluation pluralism, and lifecycle accountability across the AI lifecycle.

\begin{table*}[t]
\centering
\caption{Terminology used throughout the paper.}
\label{tab:terms}
\small
\begin{tabularx}{\textwidth}{L{0.20\textwidth}Y}
\toprule
\textbf{Term} & \textbf{Meaning in this paper} \\
\midrule
Ontology & Practical representational commitments embedded in an AI system about entities, relations, measurable attributes, evidence, and valid outputs. \\
Lifeworld & Situated background of meanings, identities, histories, and practical judgments through which people experience and organize social life. \\
Ontological flattening & Conversion of situated and contested meanings into a restricted technical category, proxy, aggregation rule, or benchmark target that is treated as neutral and difficult to contest. \\
Outcome pluralism & Pluralism at the level of outputs, answers, steering options, or represented preference distributions. \\
Procedural pluralism & Pluralism at the level of authority over problem framing, categories, evidence, aggregation, evaluation, revision, and appeal. \\
Procedural standing & Capacity of affected actors to influence, contest, revise, veto, or appeal decisions about a system, including decisions about the ontology through which the system operates. \\
Lifecycle accountability & Assignment of responsibilities, evidence, recourse, audit cadence, revision triggers, and decommissioning conditions after design and deployment. \\
\bottomrule
\end{tabularx}
\end{table*}

\section{Related Work}

Classical value pluralism rejects the reduction of legitimate values to one master scale. Berlin argues that human goods may be real and legitimate while remaining incompatible \citep{berlin1969}. Anderson and Chang show that values can be incomparable or incommensurable, making it misleading to assume that every conflict can be converted into one utility calculus \citep{anderson1993,chang2015}. For AI, the practical problem is that learning and evaluation procedures often require commensuration through labels, losses, rankings, utilities, scores, or preferences.

Science and technology studies and feminist epistemology shift the issue from value conflict to world-making. Classification systems organize work and social memory \citep{bowkerstar1999}; objects are enacted differently across practices rather than merely viewed differently from one neutral standpoint \citep{mol2002}; and objectivity requires accountable positioning rather than a view from nowhere \citep{haraway1988}. Measurement and construct validity scholarship further shows that operational choices can fail when a construct is underspecified or when a social target is treated as directly measurable \citep{jacobswallach2021measurement}. Fairness research likewise warns that abstraction can hide institutional context, background conditions, and social meaning \citep{selbst2019fairness}. These literatures imply that AI pluralism cannot be limited to different preferences over a fixed ontology. It must also ask who defines entities, relations, proxies, and evidentiary standards.

Alignment research asks how AI systems can act in accordance with human values and intentions \citep{gabriel2020,kasirzadehgabriel2023}. Reinforcement learning from human feedback, instruction tuning, and constitutional approaches have improved model behavior \citep{christiano2017deep,ouyang2022training,askell2021general,bai2022constitutional}, but they often aggregate heterogeneous judgments into one training signal. Pluralistic alignment reframes disagreement as signal. Roadmaps distinguish Overton, steerable, and distributional pluralism; social choice approaches treat feedback as collective decision-making; and projects such as STELA, PRISM, and Collective Constitutional AI show concrete ways to elicit or represent plural input \citep{sorensen2024roadmap,conitzer2024socialchoice,bergman2024stela,kirk2024prism,huang2024collective}. Our argument is not that these methods fail. The distinction is that outcome-level pluralism can still operate over fixed categories, proxies, and aggregation rules. \plg{} asks whether those choices are themselves procedurally open.

Democratic AI and procedural justice provide the institutional counterpart. Democratic AI distinguishes advisory input from processes that bind, initiate, or govern metagovernance \citep{ovadya2025democratic}. Participatory AI warns that participation becomes tokenistic when communities are consulted after core decisions are fixed \citep{arnstein1969,birhane2022power,delgado2023participatory,sloane2022participation}. Design justice identifies community-led design and accountable redistribution of design power as central to equitable systems \citep{costanzachock2020design}. Procedural justice and due process add that algorithmic legitimacy requires voice, reasons, contestation, and audit trails \citep{kinchin2024voiceless,citron2008technological,pasquale2015blackbox}. Accountability research explains why this cannot be solved at the model layer alone: AI responsibility is fragmented across model providers, data producers, deployers, and downstream institutions, and harms can arise throughout the machine-learning lifecycle \citep{widdernafus2023,suchman2007,suresh2021framework}. Documentation frameworks make disclosure more systematic, but disclosure does not by itself transfer authority \citep{gebru2021datasheets,mitchell2019modelcards}.

The contribution of ontological flattening is therefore not to restate that categories are political. That claim is established by prior work. The contribution is to specify a lifecycle failure mode for pluralistic AI: pluralistic inputs or outputs can coexist with fixed ontologies when category formation, proxy choice, aggregation, evaluation, and revision remain outside affected actors' authority.

\section{Ontological Flattening}

We define \textbf{ontological flattening} as the transformation of situated, contested, embodied, or historically specific meanings into singular or restricted categories and proxies treated as neutral descriptions of the world. Flattening occurs when disagreement is represented only as noise, non-response is treated as missingness, local categories are replaced by global labels, and a system's output is contestable but its ontology is not.

A diagnosis of flattening requires four conditions. The source concept is situated, contested, context-dependent, or historically specific. The AI system fixes that concept into a restricted category, proxy, aggregation rule, metric, or benchmark target. Disagreement, uncertainty, abstention, or context is erased, converted into noise, or made unavailable to downstream interpretation. Affected actors lack meaningful standing to contest or revise the representational commitment before or after use. These conditions distinguish flattening from ordinary abstraction. Compression is acceptable when the system's scope is limited, its rationale is documented, known alternatives and dissent are retained, and affected actors have revision or appeal rights.
\begin{figure*}[t]
\centering
\begin{tikzpicture}[
    >=Latex,
    font=\small,
    node distance=1.5cm and 1.9cm,
    box/.style={
        draw,
        rounded corners,
        fill=white,
        text width=3.7cm,
        minimum height=1.1cm,
        align=center,
        inner sep=5pt
    }
]

\node[box] (life) {Plural lifeworlds\\meaning, conflict, history};
\node[box, right=2.0cm of life] (translate) {AI translation\\schemas, labels, proxies};
\node[box, right=2.0cm of translate] (system) {System action\\predictions, scores, maps};

\node[box, above=1.25cm of translate] (contest) {Pluralistic institutions\\standing and revision};
\node[box, below=1.25cm of translate] (flat) {Ontological flattening\\category lock-in};

\draw[->, thick] 
    (life) -- node[midway, above, fill=white, inner sep=1pt] {represent} (translate);

\draw[->, thick] 
    (translate) -- node[midway, above, fill=white, inner sep=1pt] {optimize} (system);

\draw[->, thick, dashed] 
    (translate) -- (flat);

\draw[->, thick, dashed] 
    (flat.east) -| node[pos=0.78, right, fill=white, inner sep=1pt] {} (system.south);

\draw[->, thick] 
    (contest) -- node[midway, right, fill=white, inner sep=1pt] {contest} (translate);

\draw[->, thick]
    (system.north) |- node[pos=0.72, above, fill=white, inner sep=1pt] {recourse} (contest.east);

\end{tikzpicture}
\caption{AI pluralism as resistance to ontological flattening. The issue is not only inaccurate representation, but the conversion of plural lifeworlds into singular technical ontologies.}
\label{fig:flattening}
\end{figure*}
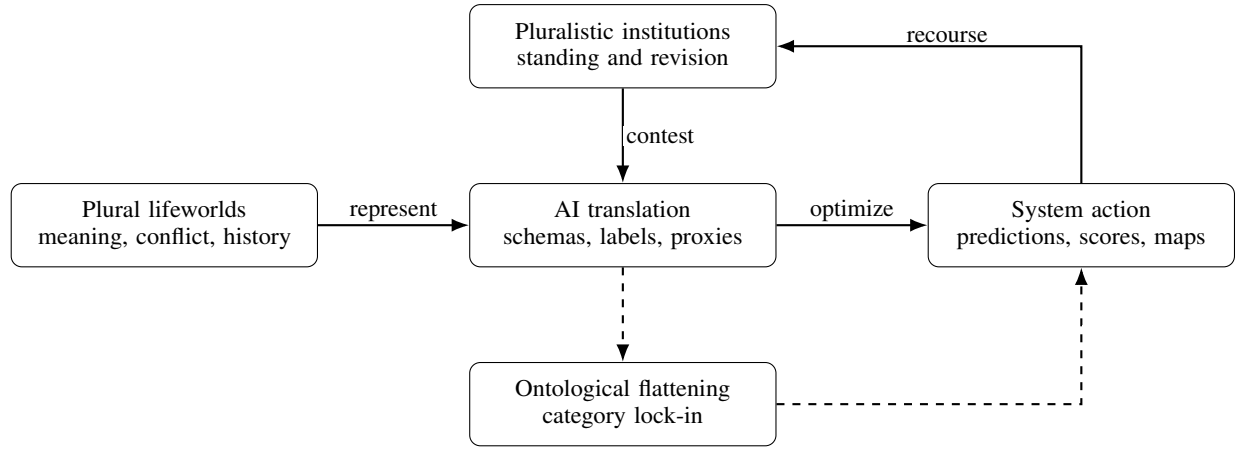

Flattening is related to bias, misclassification, measurement error, proxy failure, representational harm, and tokenistic participation, but it is not identical to them. Bias and misclassification are often diagnosed relative to an existing label space. Flattening asks whether the label space itself has unjustified authority. Measurement error concerns whether an instrument measures a specified target consistently or validly. Flattening can occur even when measurement is consistent if the target suppresses alternative meanings. Proxy failure is one mechanism of flattening, but flattening also concerns who can contest a proxy and how dissent is recorded. Representational harm concerns damaging depiction or recognition; flattening can produce representational harm, but it can also occur through apparently neutral benchmarks or maps that suppress contestable categories.

\begin{table*}[t]
\centering
\caption{Boundary between ontological flattening and adjacent concepts.}
\label{tab:boundaries}
\small
\begin{tabularx}{\textwidth}{L{0.18\textwidth}Y Y}
\toprule
\textbf{Concept} & \textbf{Primary diagnostic question} & \textbf{Relation to ontological flattening} \\
\midrule
Bias or misclassification & Are outputs wrong or uneven relative to a label space? & Flattening asks whether the label space itself wrongfully fixes a contested world. \\
Measurement or construct validity failure & Does the measure capture the intended construct? & Flattening asks whether the construct should have been singularized and who had authority to define it. \\
Proxy failure & Does a measurable proxy diverge from the target? & Proxy failure can be a mechanism of flattening when the proxy replaces situated meaning and cannot be contested. \\
Representational harm & Does the system depict or recognize a group in damaging ways? & Flattening can cause such harm, but it also applies to neutral-looking schemas, maps, and benchmark targets. \\
Tokenistic participation & Are participants consulted without power? & Tokenism is a procedural mechanism through which flattening persists despite participation. \\
Legitimate abstraction & Is compression bounded, documented, revisable, and proportionate to the task? & Abstraction is not flattening when disagreement and revision rights are preserved and the output is not presented as a complete social description. \\
\bottomrule
\end{tabularx}
\end{table*}

Technical abstraction is not automatically flattening. AI systems require compression, categorization, and operationalization. Abstraction becomes flattening when it is imposed without procedural standing for affected actors, erases known disagreement, substitutes visible proxies for situated meanings, or remains unavailable to revision after deployment. Diagnostic indicators include fixed label spaces without schema rationale, erased neutral or abstention categories, aggregation without dissent records, unavailable category appeal, and absent revision triggers.

\section{Method}

This paper is a conceptual and qualitative synthesis. It asks where pluralism fails across the AI lifecycle, how participatory AI projects preserve or compress plural values and ontologies, and what institutional mechanisms make pluralism durable beyond model outputs. The unit of analysis is the AI lifecycle: problem framing, data generation, labeling, training, evaluation, deployment, maintenance, and retirement. The paper is not a systematic review, a new controlled experiment, or a validation study for a scoring instrument. Quantitative findings are reported as source-reported results from companion case manuscripts and are used to identify mechanisms, not to estimate population-level effects.

The concept map used purposive theoretical sampling. Seed literatures were selected because they provide concepts needed to analyze pluralism in AI lifecycles: value pluralism, pluralistic alignment, democratic AI, participatory AI, procedural justice, science and technology studies, design justice, algorithmic pluralism, supply-chain accountability, and documentation. Works were included when they addressed at least one construct relevant to the framework: object of pluralism, locus of authority, evidence standard, aggregation method, lifecycle stage, or failure mode. The map was used to identify whether a source treated pluralism primarily as a value, preference, output, population, ontology, institution, or procedure.

The synthesis also incorporates 11 confidential expert interviews. Seven initial semi-structured interviews with experts in law, ethics, and AI refined construct boundaries. Four additional interviews focused on \plg{}, auditability, institutional use, and external evaluation. The interviews are used as construct-refinement evidence, not as representational inference. The paper does not report individual quotations, frequencies, or transcript-level claims. Aggregate interview themes shaped three analytic decisions: procedural authority was separated from epistemic inclusion, lifecycle accountability was treated as distinct from documentation, and auditability was defined as evidence traceability rather than numerical scoring.

The three cases were selected because they cover different lifecycle stages in a bounded empirical domain. Case A centers data formation and preference optimization for public-space generation. Case B centers participatory labels, predictive modeling, and spatial inference for streetscape inclusivity. Case C centers reliability-aware evaluation for urban vision-language models. The cases are companion manuscripts. This paper therefore treats them as illustrative evidence for mechanisms of flattening rather than independently verified empirical validation. Where the case materials do not provide enough information to resolve an interpretation, the analysis records the claim as partial, source-reported, or unknown rather than treating it as established.

\begin{table*}[t]
\centering
\caption{Evidence base and evidentiary status.}
\label{tab:datasources}
\small
\begin{tabularx}{\textwidth}{L{0.14\textwidth}L{0.22\textwidth}Y Y}
\toprule
\textbf{Source} & \textbf{Role} & \textbf{Material available to this synthesis} & \textbf{Evidence status} \\
\midrule
Concept map & Theoretical grounding & Pluralistic alignment, value pluralism, democratic AI, participatory AI, procedural justice, STS, accountability, documentation, and design justice. & Purposive map, not exhaustive review. \\
Expert interviews & Construct refinement & 11 interviews: seven initial interviews plus four \plg{}-focused interviews. & Aggregate themes only; no frequency, prevalence, or transcript-level claims. \\
Case A & Generative alignment case & Two-year participatory process; 30 community organizations; 634 initial concepts; six consolidated criteria; 37,710 pairwise comparisons across 13,462 images. & Source-reported results; used for mechanism illustration. \\
Case B & Predictive modeling case & 28 interviews; 12 focus-group raters; 20 streets; three data points per street; 60 rated locations; around 45,000 street images; 28 group-criterion outputs. & Source-reported performance; non-independence and split design remain interpretation limits. \\
Case C & Evaluation case & 100 urban scenes; 12 participants from seven organizations; 230 forms; 30 dimensions; seven vision-language models. & Source-reported benchmark; small local sample; chance baselines vary by item. \\
\bottomrule
\end{tabularx}
\end{table*}

The case analysis used a five-code matrix: ontological entry point, disagreement handling, authority and recourse, proxy limitation, and lifecycle continuation. Each case was examined across data and criteria, model and evaluation, and governance and deployment. The analysis separated four claim types: claims established by prior literature, aggregate interview-derived construct refinements, source-reported case results, and authorial inferences from the cross-case comparison. Construct validity is addressed through traceability across these claim types rather than through inter-rater statistics. \plg{} is presented as a qualitative framework and audit scaffold, not as a validated instrument with calibrated scores.

\section{Empirical Synthesis}

\textbf{Case A: pluralistic generation.} Case A asks whether community-defined, multi-criteria feedback can align text-to-image models for inclusive public spaces. Participants generated 634 initial concepts, consolidated through workshops into six locally defined criteria: Accessibility, Safety, Comfort, Invitingness, Inclusivity, and Diversity. The final dataset contains 37,710 pairwise comparisons across 13,462 images \citep{anon2025casea}. The model was fine-tuned with Direct Preference Optimization (DPO), a method that trains directly from paired preferred and dispreferred outputs \citep{rafailov2023direct}. In this case, the DPO objective encouraged transformation of multi-criteria judgments into pairwise preference targets through majority voting. The source study reports that the aligned model was favored more often than the baseline in heldout evaluation, while neutral comparisons remained common. This is not evidence that preference optimization fails. It illustrates that outcome gains can coexist with procedural compression. Neutral labels may indicate ambiguity, context-dependence, balanced trade-offs, insufficient visual evidence, or disagreement that should not be forced into a binary preference.

\textbf{Case B: participatory prediction.} Case B examines whether participatory urban perception labels can support prediction of streetscape inclusivity. The project recruited 28 interview participants and 12 focus-group raters. Participants evaluated 20 streets, with three data points per street, producing 60 rated locations. The training set expanded each location through street-level image frames, and the trained model was applied to around 45,000 street images \citep{anon2026caseb}. The source study reports internal validation and test performance, but this paper does not use the magnitude of those results as evidence of generalization because frames from the same street, nearby vantage points, or participant rating contexts could introduce non-independence if splits are not grouped by street or location. The governance issue remains even when prediction performs well internally: the model can shift inclusivity toward visible aesthetics, while participants linked inclusivity to accessibility, activity, acceptance, safety, and belonging.

\textbf{Case C: reliability-aware evaluation.} Case C asks whether vision-language models classify urban scenes in ways that align with local human annotations. It curates 100 scenes and collects annotations from 12 participants across 30 dimensions. The evaluation compares seven models under a deterministic zero-shot prompt contract \citep{anon2025casec}. The task includes single-choice and multi-label items. Accuracy is used for single-choice dimensions, Jaccard similarity for multi-label dimensions, and macro averages for dimension-level summaries. Because option counts and label structures vary, the reported macro scores from 0.16 to 0.31 cannot be interpreted against one uniform chance baseline. The source study reports that dimensions closer to visible attributes were easier than appraisal dimensions and that model performance co-varied with human reliability. Negative Krippendorff's $\alpha$ in some dimensions indicates systematic disagreement beyond chance expectations, which means those dimensions should not be treated as stable ground truth for ordinary accuracy claims. Different uses of \texttt{Not applicable} by humans and models also made abstention semantics part of the measurement result.

\begin{table*}[t]
\centering
\caption{Cross-case mechanisms and evidentiary status.}
\label{tab:findings}
\small
\begin{tabularx}{\textwidth}{L{0.14\textwidth}Y Y Y}
\toprule
\textbf{Mechanism} & \textbf{Case basis} & \textbf{Flattening risk} & \textbf{Procedural remedy} \\
\midrule
Category formation & Cases A and B began with participant concepts before consolidating criteria; Case C fixed a benchmark schema before model evaluation. & Later participation can only fill a pre-existing ontology. & Require public schema rationales, alternative label spaces, and category revision rights. \\
Disagreement and neutrality & Case A preserved neutral comparisons; Case B used group-specific ratings; Case C reported reliability and abstention. & Majority labels and averaged scores can hide ambiguity, low reliability, and minority meanings. & Report distributions, neutral rates, abstentions, reliability, aggregation sensitivity, and minority reports where appropriate. \\
Visual proxy limits & Images support spatial cues but poorly capture lived accessibility, acceptance, cultural meaning, maintenance, and belonging. & Visible form substitutes for situated meaning. & Combine visual outputs with field audits, testimony, maintenance records, and community review. \\
Lifecycle continuation & Cases document participatory inputs, but long-term authority over publication, revision, and use remains partial or unknown in the synthesis evidence. & Pluralism disappears after data collection or benchmark construction. & Specify audit cadence, recourse logs, review triggers, prohibited uses, revision rights, and decommissioning criteria. \\
\bottomrule
\end{tabularx}
\end{table*}

The cases motivate, rather than validate, a lifecycle account of pluralism. Pluralism is most fragile where a lifeworld becomes a technical object. If a system begins with fixed labels, later participation can only fill in values for those labels. Disagreement and neutrality should therefore be treated as information rather than missingness. Model gains are conditioned by aggregation: binary preferences, averaged participant scores, majority labels, thresholds, and prompt contracts are governance decisions. Outcome pluralism and procedural pluralism must therefore be evaluated separately.

\section{Pluralistic Lifecycle Governance}

We propose \methodname{} as a preliminary qualitative framework for implementing AI pluralism as a lifecycle property. \emph{Ontological openness} means that affected actors can contest categories, proxies, labels, and metrics. \emph{Epistemic inclusion} means that multiple forms of knowledge, including lived experience, are admissible evidence. \emph{Procedural authority} means that participation includes decision rights, not only advisory input. \emph{Evaluation pluralism} means that disagreement, abstention, neutrality, and reliability are reported rather than erased. \emph{Lifecycle accountability} means that pluralistic commitments persist through deployment, monitoring, revision, and retirement.

\begin{table*}[t]
\centering
\caption{Operationalizing \methodname{} as a qualitative audit scaffold.}
\label{tab:plg}
\small
\begin{tabularx}{\textwidth}{L{0.16\textwidth}Y Y Y}
\toprule
\textbf{Dimension} & \textbf{Diagnostic question} & \textbf{Sufficient evidence} & \textbf{Insufficiency indicator} \\
\midrule
Ontological openness & Can affected actors contest categories, labels, proxies, metrics, and problem definitions? & Schema rationale, contested-category log, alternatives considered, revision path, and response record. & Fixed label space with no rationale, appeal path, or record of rejected alternatives. \\
Epistemic inclusion & Which forms of knowledge count as evidence? & Recruitment rationale, compensated participation, accessibility supports, multilingual materials where needed, and inclusion of lived, technical, and institutional knowledge. & Expert-only framing, inaccessible participation, or reliance on visible proxies as complete evidence. \\
Procedural authority & Do participants have power or only voice? & Documented authority to co-define, pause, veto, escalate, appeal, or require response from a named decision-maker. & Consultation after core decisions are fixed or no obligation to respond to participant objections. \\
Evaluation pluralism & Are disagreement, neutrality, abstention, and low reliability preserved? & Distributional reporting, neutral and abstention rates, reliability, aggregation sensitivity, subgroup analysis, and minority reports where safe. & Single aggregate score or majority label that hides instability and dissent. \\
Lifecycle accountability & Who maintains pluralism after deployment or release? & Audit cadence, recourse logs, incident reporting, revision triggers, prohibited uses, versioning, and decommissioning criteria. & Responsibility ends at model release, publication, or procurement. \\
\bottomrule
\end{tabularx}
\end{table*}

\plg{} does not assign numerical scores in this paper. A minimal qualitative audit records each dimension as documented, partially documented, absent, or unknown. \emph{Documented} means that artifacts directly demonstrate both process and authority. \emph{Partially documented} means that participation or reporting exists but authority, revision, or evidence requirements remain incomplete. \emph{Absent} means that expected artifacts are not present. \emph{Unknown} means that the evidence available to the auditor is insufficient to classify the dimension. Disagreement among auditors should be recorded rather than averaged away. In institutional use, unresolved disagreement is itself evidence about governance ambiguity.

A \plg{} audit begins by defining the artifact and decision context: dataset, model, benchmark, procurement, or deployment. It then collects evidence on problem framing, data formation, label schemas, annotation instructions, aggregation rules, model objectives, evaluation reports, deployment plans, recourse procedures, and revision logs. Each dimension receives an evidence state with a supporting artifact and a limitation note. The audit report should distinguish documentation from authority. A model card may disclose a label schema, but it does not show that affected actors could revise it. A workshop may show epistemic inclusion, but it does not show procedural authority unless the workshop had binding consequences or response obligations.

Procedural authority is the dimension most likely to be overstated. Consultation gives participants an opportunity to speak. Co-design gives them influence over categories or workflows. Delegated authority gives them a defined decision right. Veto or pause rights allow them to stop a release, map publication, or use case under specified conditions. Appeal rights require a named decision-maker to respond to objections. Binding review requires that a decision cannot proceed until a defined review procedure is completed. \plg{} treats these as different evidence states because voice without response obligations can leave ontological flattening intact.

The cases illustrate diagnostic use. Case A documents community-defined concepts and criteria, but binary preference optimization leaves residual compression risk. A less compressive lifecycle design could retain criterion-specific preference heads, report neutral outcomes as a target distribution, or use multi-objective methods that preserve trade-offs. Case B documents interviews and group-specific ratings, but heatmaps create risks of proxy substitution, neighborhood stigmatization, and decontextualized planning use. A pluralistic deployment would limit map resolution, display uncertainty, prohibit punitive uses, and give community partners authority over publication and revision. Case C reports reliability, abstention, and distributional mismatch, but consensus labels can still conceal unstable semantics. A pluralistic benchmark should disclose prompt contracts, aggregation rules, label definitions, abstention semantics, and dimensions that should not be ranked.

\section{Discussion}

The synthesis explains why pluralistic outputs are insufficient. A model can produce multiple answers, steer to different users, or match a target distribution while still reproducing a fixed ontology. It may answer planning questions from several perspectives while assuming that accessibility is a stable visual property. It may generate diverse public-space images while using a training objective that collapses community disagreement. It may score well on a benchmark whose label space affected communities never authorized. Such systems are pluralistic at the output layer but monistic at the ontological layer.

Pluralism also does not require institutional paralysis. Public decisions still need to be made. Infrastructure must be built, models must be evaluated, and institutions must allocate resources. The alternative to monism is provisional consensus: decisions whose legitimacy rests on contestable procedures, public reasons, recorded dissent, and revision paths. A pluralistic AI lifecycle should preserve disagreement in the record, explain aggregation choices, and create revision triggers when deployed outputs conflict with situated experience.

Pluralism has boundaries. Some asserted ontologies deny equal standing to others, erase legally protected rights, enable targeted harm, or convert participation into harassment. A pluralistic lifecycle therefore needs boundary-setting procedures that are public, rights-respecting, and contestable. Boundary decisions should identify the excluded claim, state the rights or safety rationale, document the decision authority, preserve dissent when safe, and provide an appeal or revision path. In the present domain, a public-space model should not encode notions of safety that equate marginalized presence with risk, a generative dataset should not reward stereotypes of inclusion, and a benchmark should not treat contested social inference from pixels as a stable property of a scene.

For alignment research, the synthesis suggests three design implications. Alignment data should be treated as institutional data because labels record who had authority to define what mattered. Dataset documentation should therefore report category origin, disagreement handling, abstention semantics, and revision rights, not only demographics and collection protocols. Alignment objectives should support multi-dimensional and neutral feedback when the domain contains known trade-offs rather than treating all pluralism as a ranking problem. Benchmark scores should be reliability-aware because low human reliability can make ordinary accuracy interpretation misleading.

For governance, the urban vision cases suggest that pluralism must be tied to rights and institutions. Affected actors need standing to contest categories and not only to appeal outputs. Procurement for public-sector AI should require evidence of community involvement in problem framing and evaluation when systems operate on contested social concepts. Deployments should include revision procedures, audit cadences, recourse pathways, prohibited uses, and decommissioning triggers. These requirements must be proportionate. Participation can become burdensome when communities are asked to supply unpaid labor, when accessibility supports are absent, or when consultation fatigue accumulates. \plg{} therefore treats compensation, accessibility, multilingual materials, bounded decision scope, and response obligations as governance requirements rather than optional engagement practices.

\section{Limitations and Future Work}

The study has several limits. The empirical cases are concentrated in urban and public-space domains and rely heavily on vision-based systems. This focus makes ontological conflict observable, but it limits empirical breadth. Healthcare, education, labor, social services, and criminal justice may reveal different mechanisms. The cases were selected for thematic relevance and artifact availability rather than by an independent sampling frame. This creates a risk of confirmatory case selection, which is why the paper treats them as analytic anchors rather than validation evidence. Expert interviews are reported only in aggregate and are used for construct refinement, not prevalence claims.

The synthesis reuses results from companion projects rather than running a new controlled experiment across all lifecycle stages. The present manuscript does not provide transcript-level interview evidence, raw annotation data, full model training logs, or independent reanalysis of the companion case results. These materials would be needed for a stronger empirical paper. The current paper therefore separates source-reported case results from authorial inferences and avoids treating the cases as independent proof that \plg{} improves outcomes.

\plg{} is not yet validated as an evaluation instrument. It has no demonstrated inter-rater reliability, calibrated numerical scoring rule, or evidence that its use improves governance outcomes. Future work should apply \plg{} prospectively to external AI systems with independent raters, recorded disagreements, adjudication procedures, and outcome tracking. Additional work should test whether \plg{} changes procurement decisions, documentation quality, recourse use, or deployment revision in practice. Pluralistic procedures also require resources for workshops, accessibility supports, compensation, translation, revision, and maintenance.

\section{Conclusion}

AI pluralism is not achieved by producing many outputs, supporting many users, or collecting many preferences. Those steps matter, but they can leave intact a deeper problem: AI systems often impose one way of world-making over many others. The cases illustrate that pluralism is stronger when affected communities help define criteria, when disagreement and neutrality remain visible, when evaluation reports reliability and abstention, and when institutions allow revision. It is weakened when situated meanings are flattened into binary labels, visible proxies substitute for lived experience, or benchmark scores conceal unstable targets. This paper introduced ontological flattening as a failure mode for AI pluralism and proposed \methodname{} as a qualitative framework for making ontological contestability an explicit object of AI governance.

\bibliographystyle{icml2026}
\bibliography{pluralism_in_ai}

\end{document}